%% file: root.tex
\documentclass[letterpaper, 10 pt, conference]{ieeeconf}

\IEEEoverridecommandlockouts
\usepackage{graphicx}
\usepackage{times}
\usepackage{amsmath}
\usepackage{amssymb}
\usepackage{caption}
\usepackage{algorithm}
\usepackage{algpseudocode}
\usepackage{subcaption}
\usepackage{booktabs,multirow}
\usepackage[table]{xcolor}

\definecolor{lightblue}{rgb}{0.85, 0.93, 1.0}

\makeatletter
\let\NAT@parse\undefined
\makeatother

\usepackage[
  colorlinks=true,
  linkcolor=blue,
  citecolor=blue,
  urlcolor=blue,
  breaklinks=true
]{hyperref}

\title{\LARGE \bf
DLOSAC: Joint Spatial-Geometric Image Estimation via Delaunay-Guided Local Optimization RANSAC
}

\author{Yu Liu$^{1}$, Meiqi Zhou$^{2}$, Zipei Fan$^{1,\dagger}$, Guanyuan Feng$^{2}$, and Xuan Song$^{1}$%
\thanks{$^{\dagger}$Corresponding author ({\tt\small fanzipei@jlu.edu.cn}).}%
\thanks{$^{1}$School of Artificial Intelligence, Jilin University, China.}%
\thanks{$^{2}$School of Computer Science and Technology, Changchun University of Science and Technology, China.}%
\thanks{This research was funded by the Science and Technology Development Program of Jilin Province (No. 20250201044GX).}%
}

\begin{document}

\maketitle
\thispagestyle{empty}
\pagestyle{empty}

\begin{abstract}
Reliable image geometric estimation from noisy matching points is typically addressed using RANSAC, often combined with local optimization to refine inliers and model parameters.
However, existing paradigms are predominantly residual-centric; even heuristic attempts to enforce spatial uniformity suffer from a sequential, two-stage nature that inherently decouples distribution optimization from geometric refinement, preventing an effective trade-off.
In this paper, we propose DLOSAC, a local optimization framework that jointly incorporates both geometric consistency and spatial distribution information. We begin by constructing a Delaunay triangulation over candidate inliers to explicitly encode neighborhood topology. The geometric module introduces an edge-based consistency metric to evaluate geometric errors, while the spatial module derives distribution descriptors from triangle areas and edge attributes to quantify the clustering of inliers. Finally, we combine edge-based geometric constraints, spatial distribution regularization, and traditional point-level geometric constraints to improve the quality of geometric parameters.
Extensive experiments on six publicly available real-world datasets, covering three types of geometric estimation tasks, achieve state-of-the-art performance.
\end{abstract}

\input{sec/1_intro}
\input{sec/2_relatedwork}
\input{sec/3_problem}

\input{sec/4_method}
\input{sec/5_experiments}

\input{sec/Conclusion}

\bibliographystyle{IEEEtran}
\bibliography{ref}

\end{document}

%% file: sec/1_intro.tex
\section{Introduction}
\label{sec:intro}

    Reliable image geometric estimation from noisy matching points is a key component of many robotic perception systems, including visual localization~\cite{Visual_Loc_yan2023long,miao2024survey}, autonomous navigation~\cite{Auto_nav,zhao2024viwo}, and SLAM~\cite{zhao2024viwo,sahili2023survey,orbslam2}. The problem is typically formulated as estimating optimal motion parameters from two or more frames by solving an optimization problem~\cite{quality2_barath2020magsac++}. A standard pipeline comprises a front end and a back end: the front end extracts and matches feature points from raw sensor data (e.g., camera images), whereas the back end estimates geometric quantities (e.g., homography, fundamental and essential matrices, and relative poses). However, imaging noise and errors in feature extraction and matching often cause front-end data to deviate from ideal geometric assumptions, thereby complicating back-end estimation~\cite{quality1_barath2019magsac,sample3_barath2019progressive,sample1_matas2004robust}.

    The Random Sample Consensus (RANSAC) algorithm~\cite{fischler1981random} robustly estimates geometric model parameters under high outlier rates by repeatedly randomly sampling the minimal dataset, generating model hypotheses, and filtering out outliers based on consistency scores. Additionally, the algorithm can improve model fitting accuracy through local optimization (LO) strategies~\cite{chum2003locally}, such as point-based geometric consistency assumptions~\cite{chum2003locally,LO+RANSAC} and spatial neighborhood consistency assumptions~\cite{GC_RANSAC,PSC_RANSAC,barath2018graph,bian2017gms}, which refine the geometric error of inliers. However, despite the efficacy of LO, current paradigms remain fundamentally tethered to point-level geometric residual minimization. While bucketing-based heuristic approaches~\cite{zhang1997parameter, orbslam2, bailo2018efficient} attempt to impose spatial constraints, their sequential, two-stage nature decouples spatial distribution optimization from residual minimization, precluding an effective trade-off between the two. This motivates a central question: \textit{\textbf{How to formulate a unified optimization framework that transcends the limitations of point-based geometric verification, harmonizing the trade-off between spatial uniformity and geometric consistency.}}

   We propose DLOSAC, a local optimization framework that jointly leverages geometric consistency and spatial distribution. DLOSAC first constructs a topological graph over the inlier set via Delaunay triangulation, explicitly encoding neighborhood relations. Next, an edge-level geometric consistency metric is introduced and coupled with the conventional point-level hypothesis to refine inlier quality assessment. To capture spatial structure, distribution descriptors derived from triangle areas and edge attributes are designed to quantify inlier clustering. Finally, these cues are fused into an inlier weighting function for precise inlier selection, thereby improving the robustness and accuracy of model estimation.

    In summary, our main contributions are as follows:
    \begin{itemize}
        \item We propose DLOSAC, a novel local optimization RANSAC framework that  jointly models geometric consistency and spatial distribution to refine robust geometric estimation.
        \item We define a unified geometric consistency model, which couples a novel edge-based geometric measure with traditional point-based metrics to quantify geometric error, and a spatial distribution model to assess inlier clustering, thereby jointly improving model estimation.
        \item Extensive experiments on six public real-world datasets across three geometric estimation tasks validate that DLOSAC significantly outperforms state-of-the-art methods in both accuracy and robustness.
    \end{itemize}

%% file: sec/2_relatedwork.tex
\section{Related Work}
\label{sec:related_work}
    \subsection{Image Geometric Estimation}
        Geometric estimation is a fundamental problem in computer vision, serving as a cornerstone for key applications such as visual localization~\cite{Visual_Loc_yan2023long,miao2024survey}, autonomous navigation~\cite{Auto_nav,zhao2024viwo}, and SLAM~\cite{zhao2024viwo,sahili2023survey,orbslam2}. A typical pipeline begins with raw sensor data (e.g., camera images), from which local features are extracted and matched across views using algorithms (e.g., SIFT~\cite{sift}, SURF~\cite{surf}, ORB~\cite{rublee2011orb}) to establish a set of correspondences. Subsequently, these correspondences are used to estimate multi-view geometric parameters (e.g.,  the fundamental matrix, homography matrix, essential matrix, or relative pose) using methods like the eight-point~\cite{longuet1981computer}, normalized eight-point~\cite{hartley1997defense}, or five-point algorithms~\cite{nister2004efficient}. However, due to imaging noise and feature matching errors, this correspondence set often contains both inaccuracies (inliers with noise) and gross outliers~\cite{GC_RANSAC}. Therefore, practical systems typically employ robust estimation algorithms (e.g., RANSAC~\cite{fischler1981random} and its variants) to reject outliers, thereby ensuring a consistent and accurate estimation of the geometric parameters. 

        Moving beyond the standard RANSAC pipeline,  we aim to explore a new local optimization RANSAC framework that jointly optimizes geometric consistency and the spatial distribution of inliers to improve geometric estimation accuracy and robustness.

    \subsection{Local Optimization RANSAC}
        Random Sample Consensus (RANSAC)~\cite{fischler1981random} is primarily used to fit models to data containing a large proportion of outliers. Its core idea is to repeatedly draw minimal sample sets to generate candidate models, and then evaluate each model by the size of its inlier consensus set or by a residual metric, thereby obtaining parameter estimates that are robust to outliers. Within this framework, accuracy and efficiency can be improved along five dimensions: optimizing the sampling strategy~\cite{sample1_matas2004robust,sample2_chum2005matching,sample3_barath2019progressive}, refining model-quality evaluation~\cite{quality1_barath2019magsac,quality2_barath2020magsac++,quality3_torr2000mlesac}, designing adaptive termination criteria~\cite{chum2008optimal,quality1_barath2019magsac,sample3_barath2019progressive,sample2_chum2005matching}, introducing pre-emptive verification~\cite{chum2008optimal}, and applying local optimization~\cite{chum2003locally,LO+RANSAC,GC_RANSAC,PSC_RANSAC}. This paper primarily focuses on local optimization strategies, wherein parameters are re-estimated from inliers identified by an initial model (e.g., via nonlinear refinement or reweighted least squares) to further improve fitting accuracy~\cite{chum2003locally}. To balance robustness and efficiency, a truncated quadratic cost function can be employed, and inliers can be selectively admitted to the optimization, maintaining fit quality while reducing computation~\cite{LO+RANSAC}. Building on this idea, subsequent work incorporates spatial neighborhood consistency into inlier selection either encouraging adjacent points to share the same labels~\cite{GC_RANSAC} or to exhibit identical pixel displacement transformations~\cite{PSC_RANSAC} thereby improving outlier partitioning and enhancing the coherence and overall robustness of the inlier set.

        Prior local optimization has predominantly focused on point-level geometry. We propose an edge patch collaborative strategy that enforces geometric consistency at the edge level and models spatial distributions at the patch level, providing complementary cues that improve geometric estimation accuracy.

%% file: sec/3_problem.tex
\section{Problem Formulation}

    Given \(N\) matched keypoints \(\mathcal{M}=\{(\mathbf{x}_i,\mathbf{x}'_i)\}_{i=1}^N\), where \(\mathbf{x}_i=[x_i,y_i,1]^{\top}\in\mathbb{P}^2\) and \(\mathbf{x}'_i=[x'_i,y'_i,1]^{\top}\in\mathbb{P}^2\) are the homogeneous coordinates of corresponding points in the first (query) and second (reference) images, the goal of geometric parameter estimation is to find \(\boldsymbol{\theta}\) that best satisfies the geometric constraints induced by these correspondences. We formulate the problem as the weighted least-squares minimization:
        \begin{equation}
            \min_{\boldsymbol{\theta}} \; \sum_{i=1}^N w_i \,\big\| f(\mathbf{x}'_i,\boldsymbol{\theta},\mathbf{x}_i) \big\|_2^2 ,
            \label{eq:geom_estimation}
        \end{equation}
    where \(f(\cdot)\) denotes the constraint residual associated with the chosen geometric model. For example, if the model is the fundamental matrix \(\mathbf{F}\), then
    \(f(\mathbf{x}'_i,\mathbf{F},\mathbf{x}_i) = {\mathbf{x}'_i}^{\top}\mathbf{F}\mathbf{x}_i\) (epipolar constraint); if the model is a homography \(\mathbf{H}\), then
    \(f(\cdot)=\mathbf{x}'_i \times (\mathbf{H}\mathbf{x}_i)\) (projective constraint); and if the model is the relative pose \([\mathbf{R},\mathbf{t}]\), then
    \(f(\cdot)={\mathbf{x}'_i}^{\top}\mathbf{E}\mathbf{x}_i\) with \(\mathbf{E}=[\mathbf{t}]_{\times}\mathbf{R}\) (rigid-motion constraint). The weights \(w_i\) modulate the influence of each correspondence to improve robustness.
    
    Our goal is to introduce Delaunay triangulation to redefine the optimization weight matrix \( \mathbf{W} = \{w_i\}_{i=1}^{N} \), taking into account both geometric consistency and spatial distribution.

%% file: sec/4_method.tex
\section{Methodology}
\begin{figure*}[t!]
    \centering
    \includegraphics[width=1.0\linewidth]{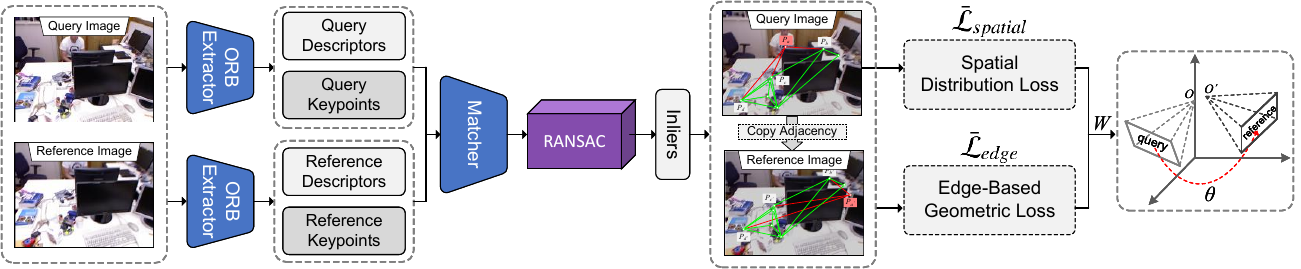}
    \caption{Overview of our framework. Given RGB images from different viewpoints, initial matches are obtained through ORB feature extraction and refined using RANSAC. A Delaunay triangulation is then built over the inliers to capture neighborhood relationships, followed by the computation of spatial distribution and geometric consistency losses. Finally, a weighted matrix is used to optimize the geometric parameters.}
    \label{fig:framework}
\end{figure*}

    \subsection{Preliminaries}
        \subsubsection{Delaunay triangulation}
            Delaunay triangulation (DT) is a widely used spatial discretization method that partitions a given planar point set \( \mathcal{P} = \{\mathbf{p}_i\}_{i=1}^n \subset \mathbb{R}^2 \) into a collection of non-overlapping triangular elements. Its fundamental principle is the \textit{empty circumcircle property}: for any triangle in the triangulation, the circumcircle contains no other points from the set except the triangle’s three vertices~\cite{preparata2012computational}. 

            DT maximizes the minimum angle among all triangles, effectively preventing the formation of highly acute or degenerate elements and ensuring a more uniform and numerically stable mesh. Moreover, Delaunay triangulation naturally induces a graph structure \( \mathcal{G} = (\mathcal{V}, \mathcal{E}) \), where \( \mathcal{V} = \mathcal{P} \) denotes the set of nodes and \( \mathcal{E} = \{(\mathbf{p}_i, \mathbf{p}_j) \mid \mathbf{p}_i, \mathbf{p}_j \in \mathcal{P}\} \) defines the adjacency relationships among them. This geometry-induced graph captures the intrinsic local topology of the point set and has been widely applied in medical imaging, computer vision, and robotic downstream tasks~\cite{elshakhs2024comprehensive}.

        \subsubsection{Local Optimization RANSAC}

            Image geometric estimation seeks to recover scene or camera motion parameters from feature correspondences across multiple views. Let \(\mathcal{M}=\{(\mathbf{x}_i,\mathbf{x}'_i)\}_{i=1}^{N}\) denote the set of putative correspondences, \(\boldsymbol{\theta}\) the model parameters (e.g., a fundamental or essential matrix), and \(\varepsilon(\mathbf{x}_i,\mathbf{x}'_i;\boldsymbol{\theta})\) a geometric residual (e.g., the symmetric epipolar distance). Given an inlier threshold \(\tau>0\), a correspondence \((\mathbf{x}_i,\mathbf{x}'_i)\) is considered an inlier if \(\varepsilon(\mathbf{x}_i,\mathbf{x}'_i;\boldsymbol{\theta})<\tau\). RANSAC~\cite{fischler1981random} formulates estimation as a maximum-consensus (maximum inlier count) problem:
            \begin{equation}
              \hat{\boldsymbol{\theta}}
              = \underset{\boldsymbol{\theta} \in \Theta}{\arg\max}\; I(\boldsymbol{\theta}),
              \quad
              I(\boldsymbol{\boldsymbol{\theta}})
              = \sum_{i=1}^{N} \mathbf{1}\!\left[\varepsilon\!\bigl(\mathbf{x}_i,\mathbf{x}_i';\boldsymbol{\theta}\bigr) < \tau\,\right],
            \end{equation}
            where \(\Theta\) is the feasible parameter space and \(\mathbf{1}[\cdot]\) denotes the indicator function, which equals \(1\) if the condition holds and \(0\) otherwise.

            The algorithm proceeds iteratively by (i) sampling a minimal subset from \(\mathcal{M}\) to generate a hypothesis \(\boldsymbol{\theta}_k\), (ii) scoring \(\boldsymbol{\theta}_k\) by evaluating \( I(\boldsymbol{\theta}_k)\) over all matches, and (iii) retaining the highest-scoring hypothesis; iterations terminate once a confidence-driven stopping criterion is met, at which point the hypothesis with the largest consensus set is selected. To further enhance accuracy and stability, LO-RANSAC~\cite{chum2003locally} then performs localized optimization when a sufficiently large consensus set has been identified: letting \(\boldsymbol{\theta}^*\) denote the current best model with inliers \(I^{*}\), the parameters are re-estimated from all inliers (or a filtered subset \(\tilde{I}^{*} \subseteq I^{*}\)), which reduces threshold \( \tau\) induced bias, improves numerical conditioning, and delivers superior performance under measurement noise and non-uniform outlier distributions. 

    \subsection{Proposed Local Optimization RANSAC}
        \subsubsection{Edge-Based Geometric Consistency Loss}
            \begin{figure}[t]
                \centering
                \begin{subfigure}[t]{0.461\linewidth}
                    \centering
                    \includegraphics[width=\linewidth]{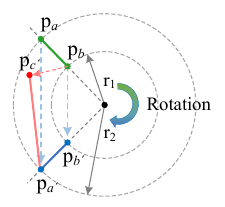}
                    \caption{}
                    \label{fig:etcm:a}
                \end{subfigure}\hfill
                \begin{subfigure}[t]{0.538\linewidth}
                    \centering
                    \includegraphics[width=\linewidth]{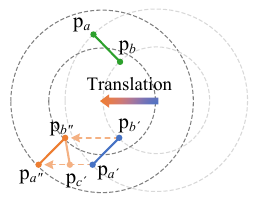}
                    \caption{}
                    \label{fig:etcm:b}
                \end{subfigure}
                \caption{Edge-based geometric consistency models: (a) rotation and (b) translation.}
                \label{fig:etcm}
            \end{figure}

            For clarity, we first analyze edge-based geometric consistency under a simplified camera motion model and then generalize the conclusions to general camera movement. As shown in Fig.~\ref{fig:etcm}(\subref{fig:etcm:a}), the matched points \( \mathbf{p}_a \) and \( \mathbf{p}_b \) in the query image are transformed into the corresponding points \( \mathbf{p}_{a'} \) and \( \mathbf{p}_{b'} \) in the reference image after the camera undergoes rotation. The adjacency relationships in the query image (such as \( \mathbf{p}_a \) and \( \mathbf{p}_b \)) are represented by the line segment \(\ell(\mathbf{p}_a,\mathbf{p}_b)\) via Delaunay triangulation, while the adjacency relationships in the reference image (such as \( \ell(\mathbf{p}_{a'},\mathbf{p}_{b'}) \)) are determined through point correspondences between the query and reference images. From the figure, it can be observed that after the camera undergoes a rotational transformation, the lengths of the adjacency segments \(\ell(\mathbf{p}_a,\mathbf{p}_b) \) and \( \ell(\mathbf{p}_{a'},\mathbf{p}_{b'}) \) remain consistent. However, when an incorrect match occurs, such as when \( \mathbf{p}_b \) is erroneously matched to \( \mathbf{p}_{c'} \), the distance between adjacent points (such as \( \ell(\mathbf{p}_a,\mathbf{p}_b) \) and \( \ell(\mathbf{p}_{a'},\mathbf{p}_{c'}) \)) will deviate. Accordingly, we introduce a regularization constraint: under pure rotation, edges connecting neighboring points produced by correct correspondences exhibit smaller length variations between the query and reference images than those produced by mismatches.

            Fig.~\ref{fig:etcm}(\subref{fig:etcm:b}) illustrates the correspondences between an adjacent feature point pair, $\mathbf{p}_{a'}$ and $\mathbf{p}_{b^{'}}$, in the query image and their respective matches, $\mathbf{p}_{a''}$ and $\mathbf{p}_{b''}$, in the reference image following camera translation. According to the principle of triangulation~\cite{hartley1997triangulation}, an object's depth is inversely proportional to its parallax. Specifically, when the depth difference between neighboring points $\mathbf{p}_{a'}$ and $\mathbf{p}_{b'}$ in the query image is small, their pixel displacements across viewpoints are also small. Consequently, the connecting edges $\ell(\mathbf{p}_{a'},\mathbf{p}_{b'})$ and $\ell(\mathbf{p}_{a''},\mathbf{p}_{b''})$ in the query and reference images, respectively, are more likely to maintain consistent lengths. Conversely, a large depth difference may lead to correspondences such as ($\mathbf{p}_{a'}\rightarrow{\mathbf{p}}_{c'}$) and ($\mathbf{p}_{b'}\rightarrow{\mathbf{p}}_{b''}$). Even when these correspondences are correct, the edge lengths $\ell(\mathbf{p}_{a'},\mathbf{p}_{b'})$ and $\ell(\mathbf{p}_{c'},\mathbf{p}_{b''})$ can still be inconsistent between the query and reference images. Based on the preceding analysis, we define the edge-based geometric consistency loss ($\mathcal{L}_{\text{edge}}$) as:
            \begin{align}
                \mathcal{L}_{\text{edge}}&= (1- d_{ab})\cdot (\ell(\mathbf{p}_{a'},\mathbf{p}_{b'}) - \ell(\mathbf{p}_{a''},\mathbf{p}_{b''}))^{2},
                \label{eq: loss_edge}
            \end{align}
            where $d_{ab}$ denotes the pixel distance between neighboring points $\mathbf{p}_{a'}$ and $\mathbf{p}_{b'}$, normalized by the length of the image diagonal. Inspired by~\cite{PSC_RANSAC}, we introduce the weight $(1-d_{ab})$: points that are closer in the image (small $d_{ab}$) are more likely to have similar depths and thus to satisfy the geometric consistency constraint on adjacent edges. Conversely, points that are farther apart (large $d_{ab}$) are more likely to exhibit significant depth differences, increasing the probability that the edge connecting them becomes inconsistent across views. This assumption extends to general camera models, as distant points are more susceptible to perspective-induced geometric distortions and therefore deviate more from the imposed constraint.

\begin{figure}
                \centering
                
                \includegraphics[width=0.99\linewidth]{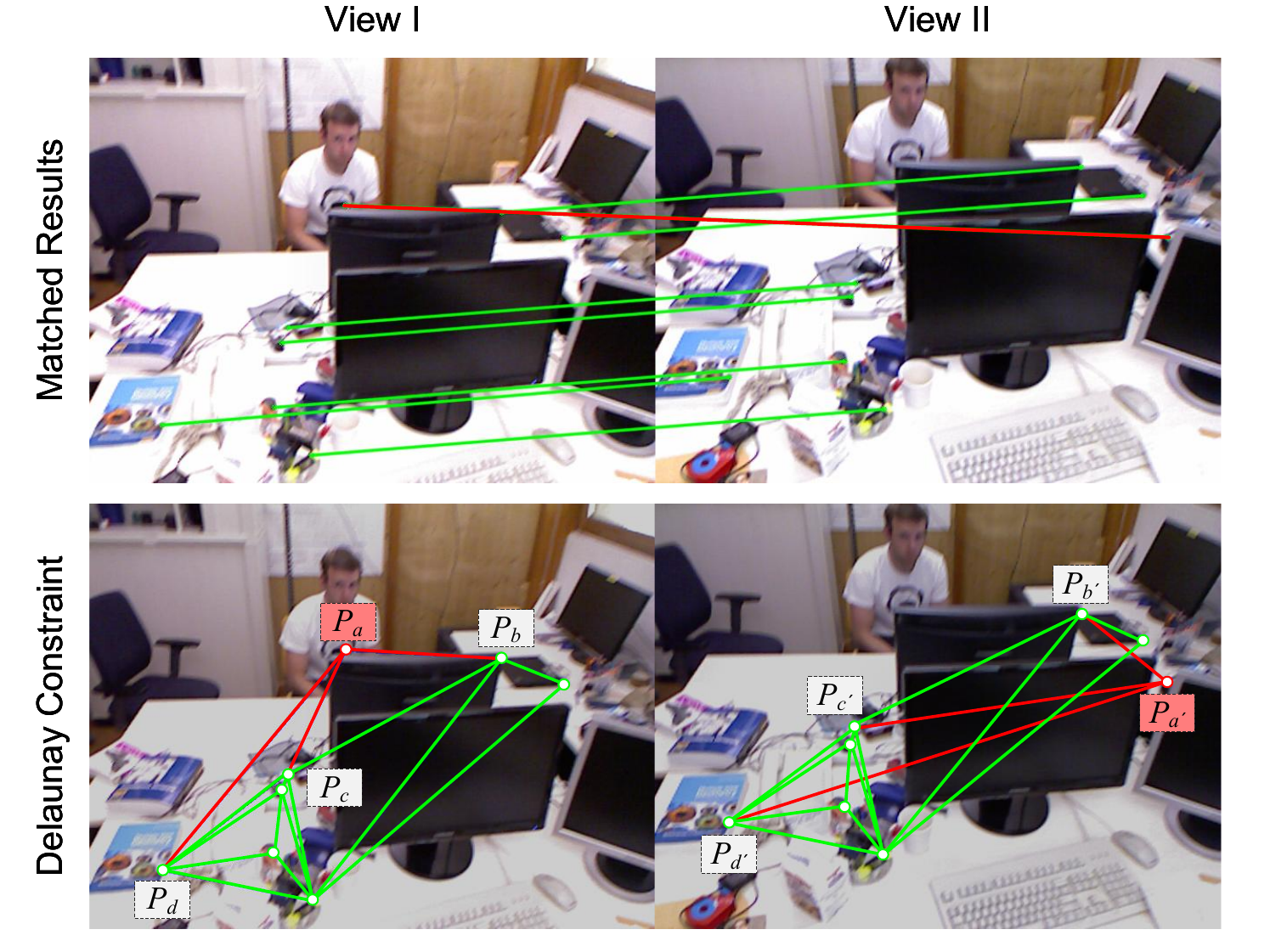}
                \caption{Illustration of an instance of edge-based geometric consistency.}
                \vspace{-1em}
                \label{fig:Delaunay_constraint}
            \end{figure}

            Note that the $ \mathcal{L}_{\text{edge}}$ is a discriminative prior derived under a simplified camera model rather than a physical law. In practice, mismatches often result in substantial pixel-level deviations. Consequently, even under general camera motion model, this prior is also more likely to be violated (see Fig.~\ref{fig:Delaunay_constraint}). More importantly, we use $ \mathcal{L}_{\text{edge}}$ solely as a regularizer (details in Sec.~\ref{sec:DGLO}), while the primary constraints are imposed by point-based geometric errors (e.g., epipolar constraint or homography residuals).

        \subsubsection{Spatial Distribution Loss}

            As shown in Fig.~\ref{fig:Delaunay_constraint}, after Delaunay triangulation, each matched point serves as a shared vertex for multiple adjacent triangles. To quantify each point's spatial distribution, we utilize the areas and largest internal angles of its adjacent triangles. These geometric properties are then used to define the Area Descriptor ($D_A^{(i)}$) and the Shape Descriptor ($D_S^{(i)}$) as follows:

                \begin{align}
                D_A^{(i)} &= \sqrt{ \frac{1}{m_i - 1} \sum_{j=1}^{m_i} \left( \frac{A(T_{ij}) \cdot n}{\sum_{k=1}^{n} A(T_k)} - 1 \right)^2 } \label{eq:area_descriptor}, \\
                D_S^{(i)} &= \sqrt{\frac{1}{m_i - 1} \sum_{j=1}^{m_i} \left(\frac{3 \cdot \max(J_{ij})}{\pi} - 1 \right)^2 }, \label{eq:shape_descriptor}
                \end{align}
            where $m_i$ is the number of triangles adjacent to the $i$-th matched keypoint, and $n$ is the total triangle count in the triangulation. $A(T_{ij})$ denotes the area of the $j$-th triangle adjacent to keypoint $i$, $\sum_{k=1}^{n} A(T_k)$ is the total area of all $n$ triangles, and $\max(J_{ij})$ is the largest internal angle (in radians) of triangle $T_{ij}$.

            Smaller values for the Area Descriptor ($D_A^{(i)}$) and the Shape Descriptor ($D_S^{(i)}$) indicate that the point's adjacent triangles are closer to equilateral and their areas are nearer the global mean, respectively. This implies a more uniform spatial distribution for the matched point. Since clustered points can introduce local errors, we aim to promote uniformity by minimizing these descriptors. We thus define the spatial distribution loss for the $i$-th point, $L^{(i)}$, as:
            \begin{align}
            \mathcal{L}_{\text{spatial}}^{(i)} = D_A^{(i)} \cdot D_S^{(i)}. 
            \label{eq:spatial_loss}
            \end{align}

        \subsubsection{Delaunay-Guided Local Optimization}   
        \label{sec:DGLO}
            An overview of the proposed Delaunay-Guided Local Optimization RANSAC (DLOSAC) framework is shown in Fig.~\ref{fig:framework}. The pipeline begins by constructing a Delaunay triangulation on the inlier set $\mathcal{P}_{\text{inlier}} \in \mathcal{M}$, thereby establishing topological neighborhood relationships among the matches. Subsequently, for each node (keypoint) $\mathbf{p} \in \mathcal{P}_{\text{inlier}}$ within the triangulation, we define two structured losses: (i) the mean edge geometric consistency loss $\bar{\mathcal{L}}_{\text{edge}}$ over all its adjacent edges (Eq.~\ref{eq: loss_edge}), and (ii) the mean spatial distribution loss $\bar{\mathcal{L}}_{\text{spatial}}$ over all its incident triangles (Eq.~\ref{eq:spatial_loss}).

            To derive a robust weight $w_p$ for each node, we first compute a unified loss $\mathcal{L}_p = \bar{\mathcal{L}}_{\text{edge}} + \lambda_{\text{spatial}} \bar{\mathcal{L}}_{\text{spatial}}$, where the $\lambda_{spatial}$ coefficients act as balancing factors. This unified loss is then transformed into a weight via an exponential decay function, $w_{p} = \exp(-\mathcal{L}_p / \sigma^2)$, where $\sigma^2$ is a scaling parameter. This weighting scheme ensures that nodes with higher geometric consistency and better spatial distribution (i.e., lower $\mathcal{L}_p$) receive exponentially higher weights, while outliers or poorly-structured nodes are effectively down-weighted. Finally, these node-specific weights $w_p$ are employed to solve for the geometric parameters $\boldsymbol{\theta}$ using the Weighted Least-Squares (WLS) formulation (Eq.~\ref{eq:geom_estimation}).

        \begin{algorithm}[t!]
        \caption{Delaunay-Guided Local Optimization}
        \label{alg:dglo}
        \begin{algorithmic}[1]
        \Require
            Initial match set $\mathcal{M}$; 
            Balancing factor $\lambda_{\text{spatial}}$;
            Scaling parameter $\sigma^2$
        \Ensure
            Estimated geometric parameters $\boldsymbol{\theta}$
        
        \State $\mathcal{P}_{\text{inlier}} \leftarrow \text{RANSAC}(\mathcal{M})$
        \State $\mathcal{G}(\mathcal{V}, \mathcal{E}) \leftarrow \text{DelaunayTriangulation}(\mathcal{P}_{\text{inlier}})$
        \State $W \leftarrow \emptyset$ \Comment{Initialize an empty set for weights}
        
        \For{each node $\mathbf{p} \in \mathcal{V}$}
            \State $\bar{\mathcal{L}}_{\text{edge}} \leftarrow \text{MeanEdgeLoss}(\mathbf{p}, \mathcal{G})$ \Comment{per Eq.~(\ref{eq: loss_edge})}
            \State $\bar{\mathcal{L}}_{\text{spatial}} \leftarrow \text{MeanSpatialLoss}(\mathbf{p}, \mathcal{G})$ \Comment{per Eq.~(\ref{eq:spatial_loss})}
            
            \State $\mathcal{L}_p \leftarrow \bar{\mathcal{L}}_{\text{edge}} + \lambda_{\text{spatial}} \bar{{\mathcal{L}}}_{\text{spatial}}$ 
            \State $w_p \leftarrow \exp(-\mathcal{L}_p / \sigma^2)$  
            \State $\mathbf{W} \leftarrow \mathbf{W} \cup \{w_p\}$
        \EndFor
        
        \State $\boldsymbol{\theta} \leftarrow \text{WLS}(\mathcal{V}, \mathbf{W})$ \Comment{per Eq.~(\ref{eq:geom_estimation})}
        \State \Return $\boldsymbol{\theta}$
        \end{algorithmic}
        \end{algorithm}

%% file: sec/5_experiments.tex
\section{Experiments}
    The experimental section is organized to demonstrate the effectiveness and robustness of our method. Specifically, Sec.~\ref{sec:Comparison} provides a comparative study against state-of-the-art geometric estimation techniques. Sec.~\ref{sec:ablation} presents an ablation study to quantify the benefits of the edge–geometry consistency and spatial-distribution losses. Sec.~\ref{sec:sensitivity} evaluates the influence of critical hyperparameters on the final estimation quality.
    
    \subsection{Experimental Setup}

\begin{figure}[t!]
            \centering
            
            \begin{subfigure}[t]{0.95\linewidth}
                \centering
                \includegraphics[width=0.325\linewidth]{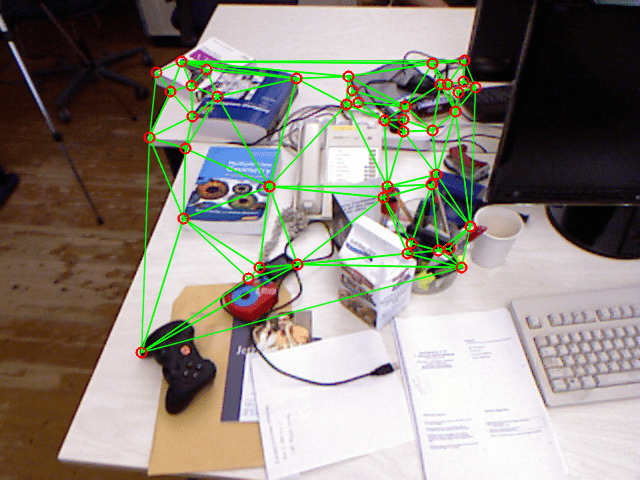}
                \includegraphics[width=0.65\linewidth]{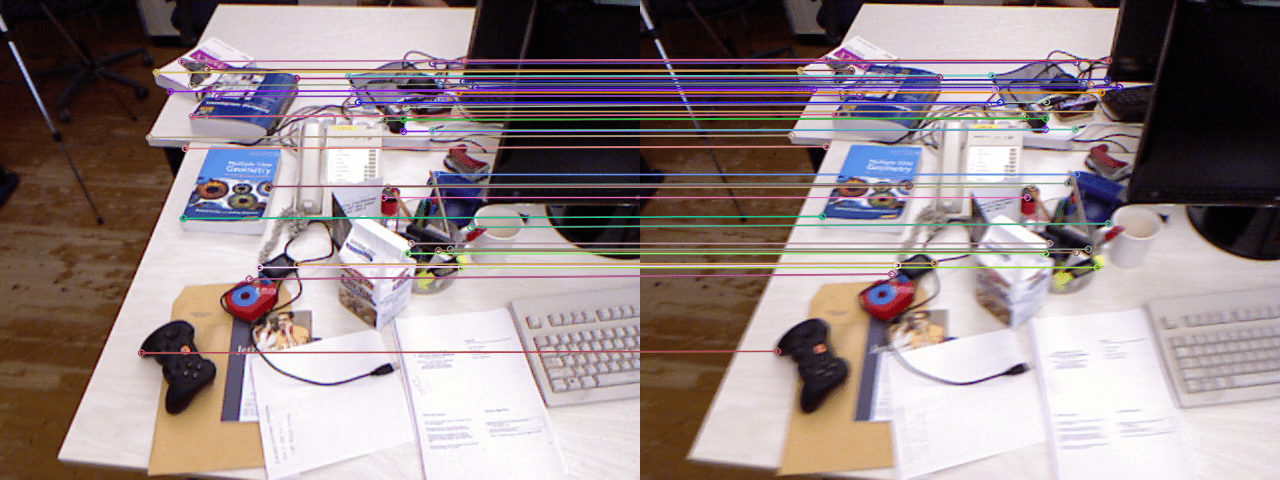}
                
                \caption{}
             \end{subfigure}

             \begin{subfigure}[t]{0.945\linewidth}
                \centering
                \includegraphics[width=\linewidth]{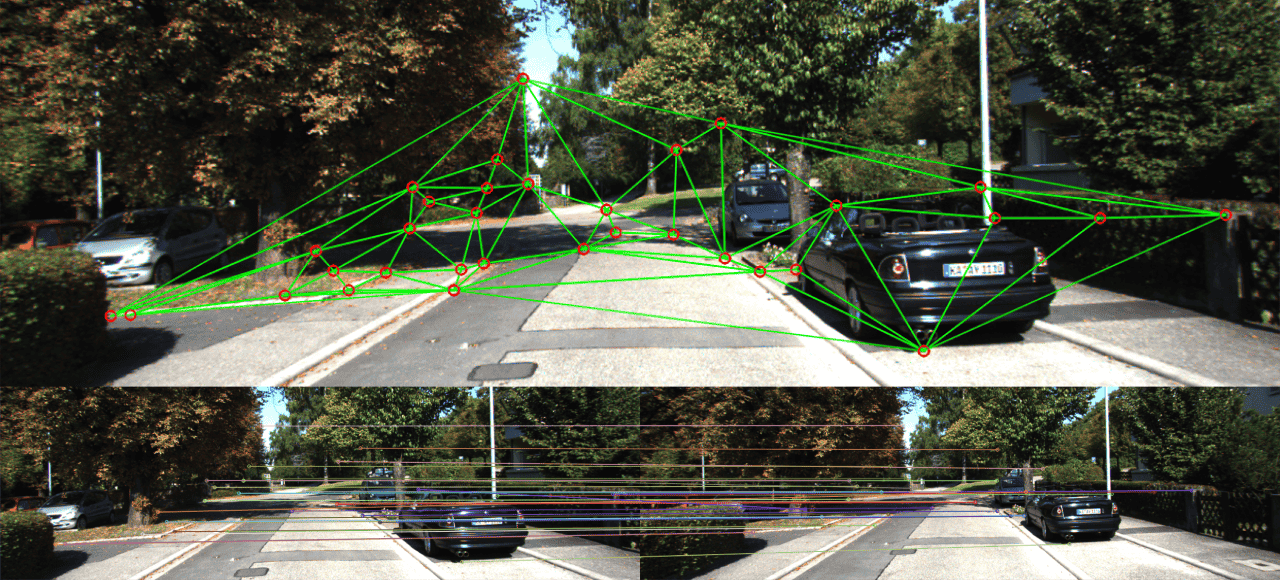}
                \caption{}
             \end{subfigure}

             \begin{subfigure}[t]{0.95\linewidth}
                \centering
                \includegraphics[width=0.325\linewidth]{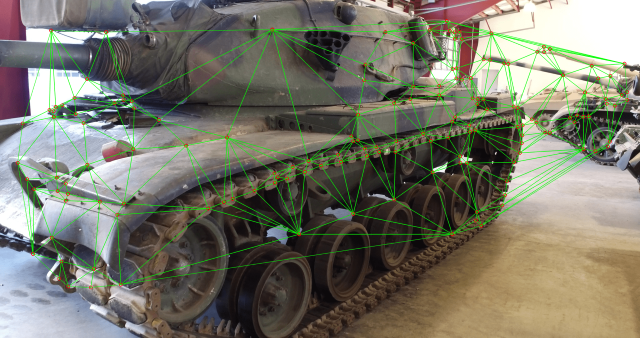}
                \includegraphics[width=0.65\linewidth]{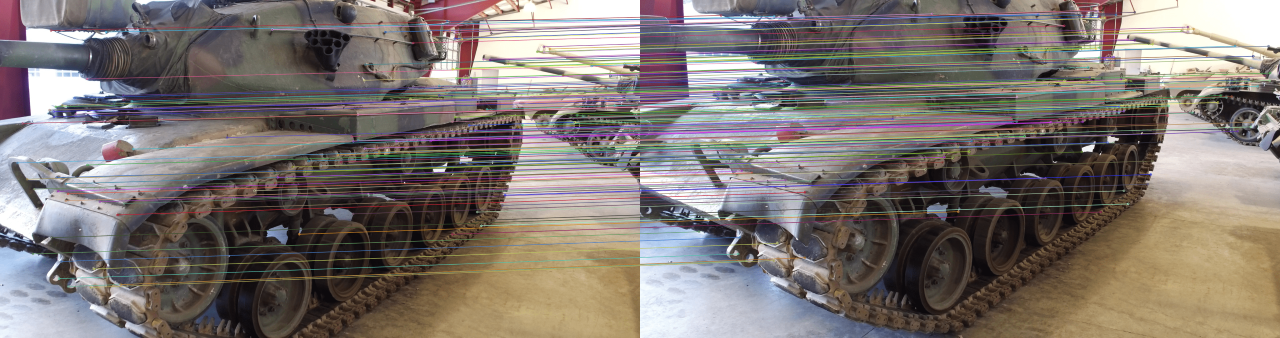}
                
                \caption{}
             \end{subfigure}

             \begin{subfigure}[t]{0.95\linewidth}
                \centering
                \includegraphics[width=0.325\linewidth]{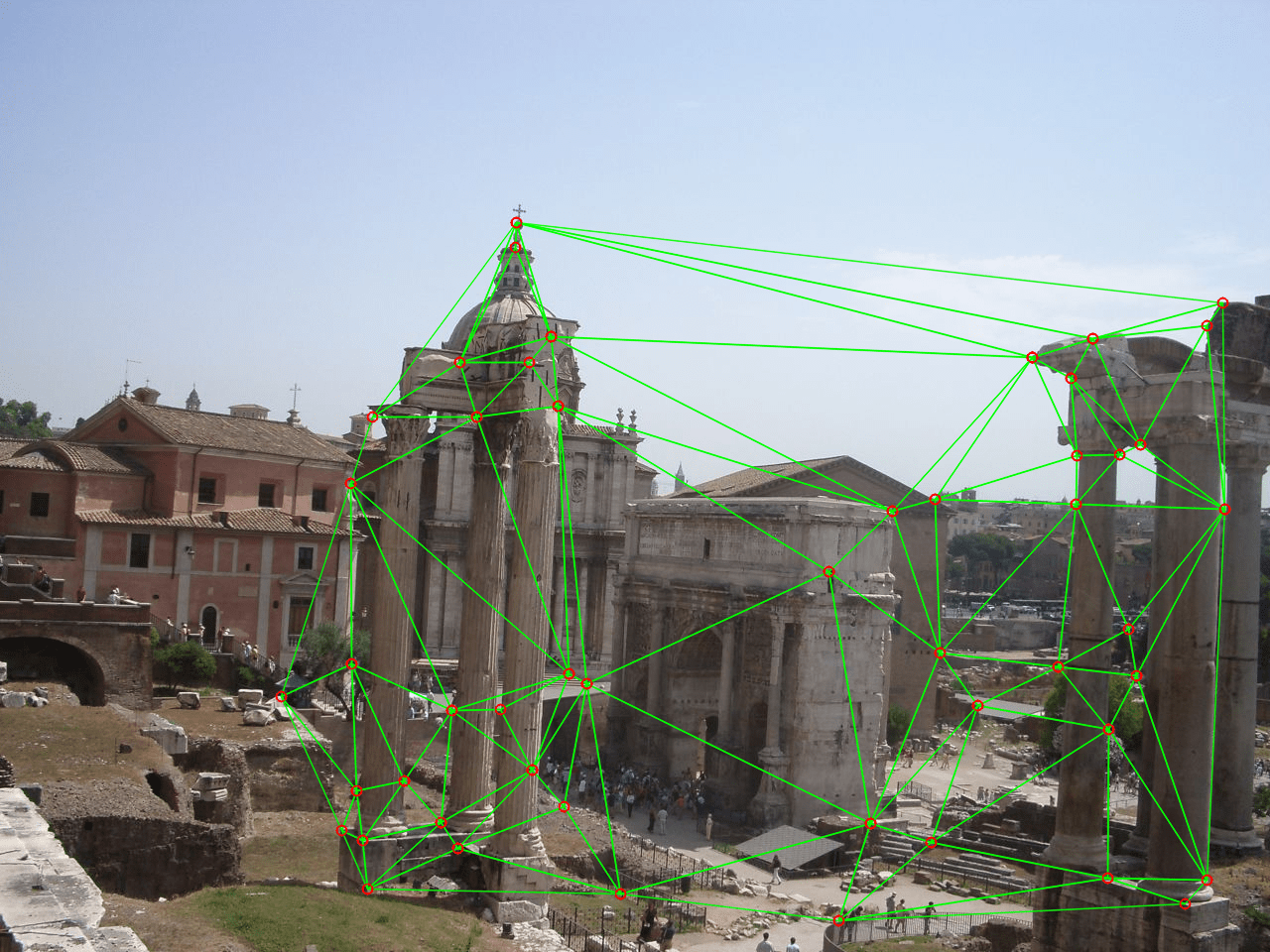}
                \includegraphics[width=0.65\linewidth]{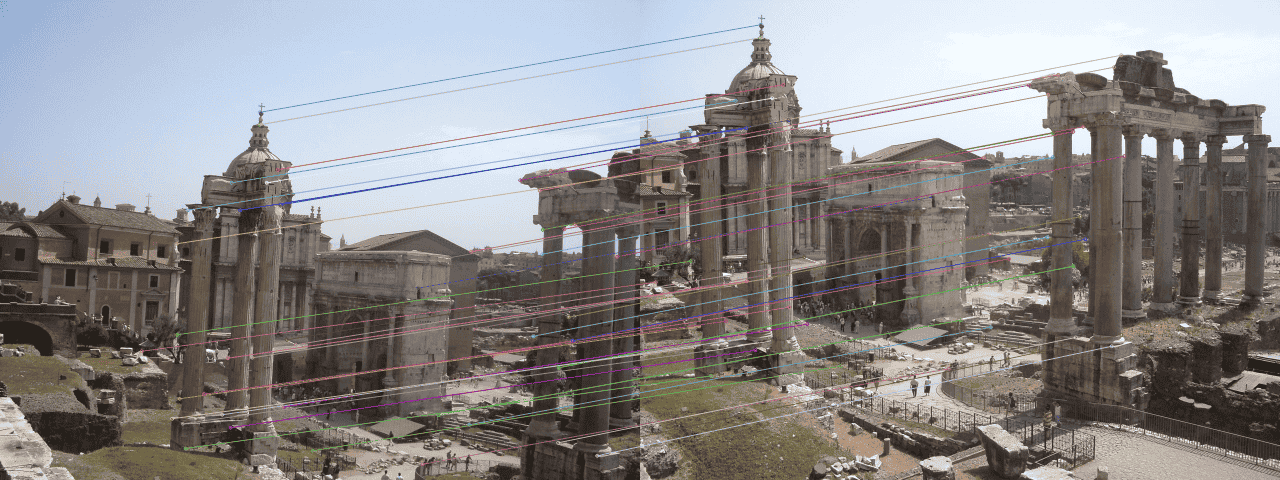}
                
                \caption{}
             \end{subfigure}

             \begin{subfigure}[t]{0.95\linewidth}
                \centering
                \includegraphics[width=0.325\linewidth]{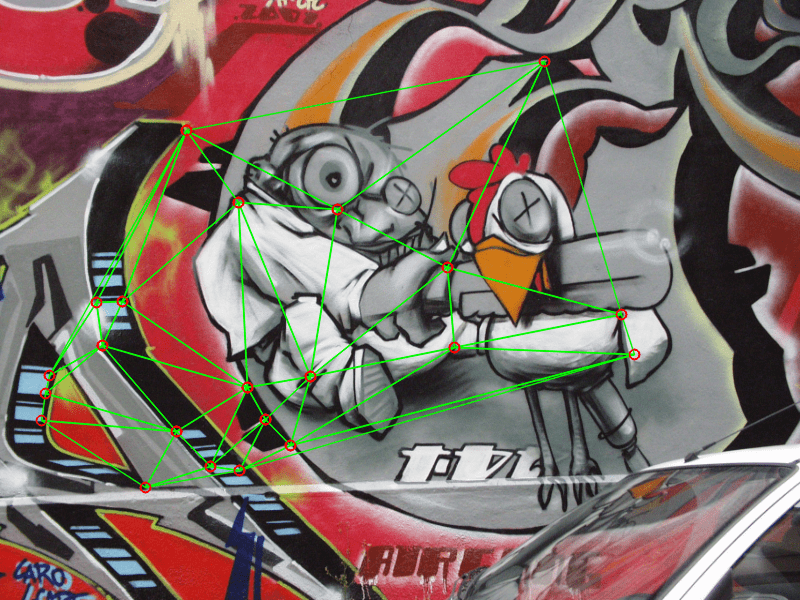}
                \includegraphics[width=0.65\linewidth]{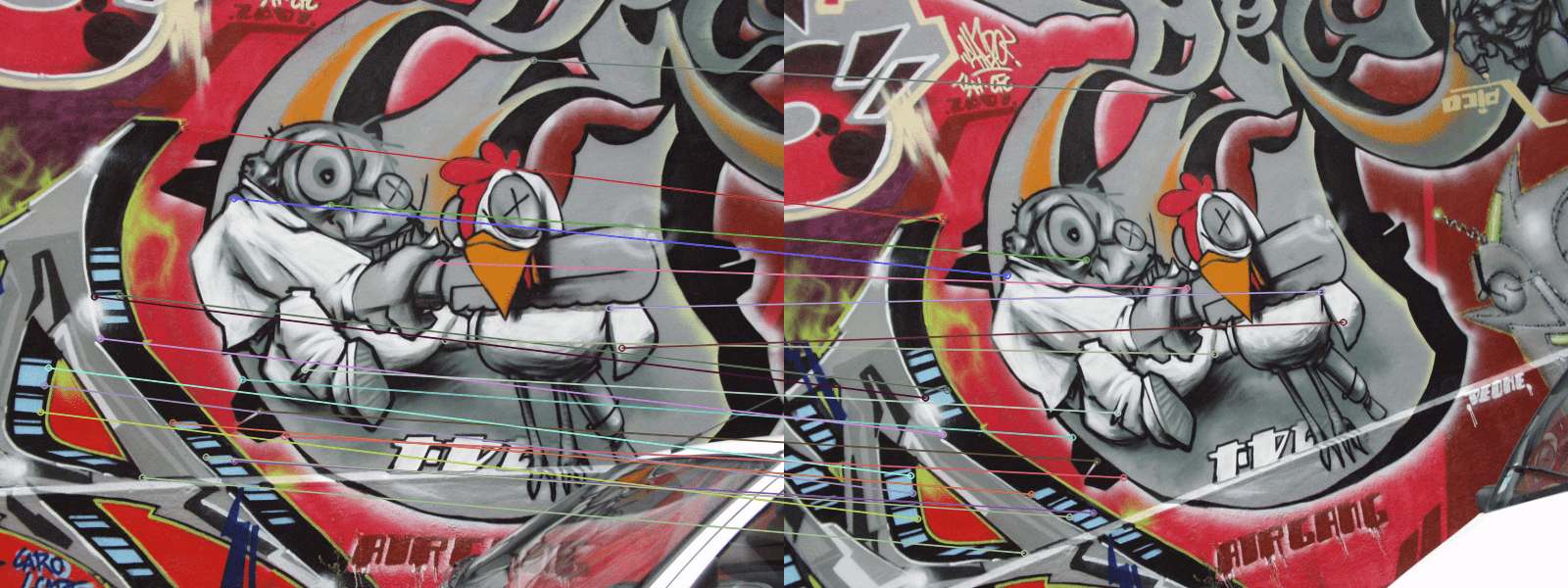}
                
                \caption{}
             \end{subfigure}

            \caption{Delaunay triangulation results and corresponding matching instances for image evaluation benchmarks: (a) TUM, (b) KITTI, (c) T\&T, (d) CPC, (e) EVD.}
            \label{fig:dataset}
        \end{figure}

        \textit{1)~Benchmark:} Following the benchmark of \cite{GC_RANSAC}, we evaluate three tasks: fundamental matrix estimation, essential matrix (relative pose) estimation, and homography estimation. For the first two tasks, we conduct experiments on four real-world datasets: TUM~\cite{TUM}, KITTI~\cite{KITTI}, Tanks and Temples (T\&T)~\cite{T&T}, and Community Photo Collection (CPC)~\cite{CPC}. TUM consists of indoor video sequences with a frame resolution of \(640\times480\) pixels; KITTI comprises consecutive frames captured by a vehicle-mounted camera with an image resolution of \(1226\times370\) pixels. In both datasets, most image pairs have short baselines, making epipolar-geometry estimation relatively easy, with a high inlier ratio. T\&T provides real-object images for reconstruction, mostly wide-baseline pairs, with resolutions ranging from \(1080\times1920\) to \(1080\times2048\) pixels. CPC collects landmark photos from Flickr with varying resolutions and is likewise dominated by wide-baseline pairs. For homography estimation, we use two datasets, Homo (16 image pairs) and EVD (15 image pairs)~\cite{LO+RANSAC}. They provide point correspondences, and image sizes span from \(329\times278\) to \(1712\times1712\) pixels. To construct the evaluation benchmark, we uniformly sample 1,000 image pairs from each dataset with more than 1,000 images, as shown in Fig.~\ref{fig:dataset}.

        \textit{2)~Baselines:} \textbf{RANSAC}~\cite{fischler1981random}, with local optimization disabled, is used as the baseline to evaluate the accuracy and stability of robust estimation. \textbf{LO-RANSAC}~\cite{chum2003locally} performs local optimization (e.g., iterative least squares) on the inlier set of the current best model to refine both the model parameters and the inlier membership. \textbf{LO$^{+}$-RANSAC}~\cite{LO+RANSAC} extends the LO stage with a truncated-quadratic loss and an upper bound on the inlier count. \textbf{GC-RANSAC}~\cite{GC_RANSAC} employs graph-cut energy minimization to jointly exploit residual magnitudes and spatial coherence for inlier/outlier labeling and subsequent model refinement. \textbf{PSC-RANSAC}~\cite{PSC_RANSAC} clusters correspondences in displacement space to remove mismatches, followed by model refinement.

        \textit{3) Evaluation Metrics:} We evaluate the accuracy of \textit{fundamental matrix} and \textit{homography} estimation using the mean ${\epsilon_{\text{avg}}}$ (in pixel) and standard deviation $\epsilon_{\mathrm{std}}$ (in pixel) of the pixel error. The improvement rate $f_{\mathrm{imp}}$ quantifies the performance of a competing method relative to ours and is defined as:
                \begin{equation}
                f_{\mathrm{imp}}(\epsilon_{\mathrm{avg}})=\frac{\epsilon_{\text{avg}}^{\text{baseline}}-{\epsilon}^{\text{ours}}_{\text{avg}}}{\epsilon_{\text{avg}}^{\text{baseline}}}\times 100\% .
                \end{equation}
                
            For relative pose estimation, we report the rotation error $\epsilon_{\text{R}}$~(in degree) and the translation error $\epsilon_{\text{t}}$~(in degree), and define $f_{\mathrm{imp}}(\epsilon_{\mathrm{avg}})$, where $\epsilon_{\mathrm{avg}}=\tfrac{1}{2}\bigl({\epsilon}_{\text{R}}+{\epsilon}_{\text{t}}\bigr)$, to characterize the overall error level.

    \subsection{Comparison with state-of-the-art methods}
    \label{sec:Comparison}
        \subsubsection{Quantitative Geometry Estimation Evaluation}
        \input{table/table1}

        Table~\ref{tab:t1} summarizes the results of three image-geometry estimation tasks (1st column) evaluated on six benchmark datasets and compares them with the strongest baselines (1st row). For each task, we additionally report the cross-dataset average (Avg.). Under a unified evaluation protocol, DLOSAC achieves state-of-the-art performance across all three tasks, improving over the strongest baseline to date by (+12.85\%) on fundamental-matrix estimation, (+13.85\%) on relative-pose estimation, and (+15.88\%) on homography estimation.
        
       The performance of each method is analyzed as follows: RANSAC~\cite{fischler1981random} lacks local optimization and is therefore highly sensitive to threshold selection and noise, yielding the weakest overall performance. LO-RANSAC~\cite{chum2003locally} and $\text{LO}^{+}$-RANSAC~\cite{LO+RANSAC} introduce local optimization on point-wise geometric errors; the latter further refines the inlier set through a more efficient sampling strategy, thereby improving model fit. However, these local refinements and repeated evaluations increase computational cost, and the methods can become trapped in suboptimal solutions when inliers are unevenly distributed or local minima are present. As strong baselines, GC-RANSAC~\cite{GC_RANSAC} and PSC-RANSAC~\cite{PSC_RANSAC} explicitly incorporate a neighborhood consistency prior in addition to point-wise geometric errors where nearby points tend to share inlier/outlier labels or exhibit similar pixel displacements, which improves accuracy and robustness. By contrast, DLOSAC explicitly models both spatial geometric consistency and the distribution of correspondences during local optimization, coupling them with the standard point-wise geometric constraint. This enhances the geometric consistency of correspondences while regularizing their spatial distribution to mitigate cluster-induced overfitting. Notably, T\&T~\cite{T&T} and CPC~\cite{CPC} are long-baseline datasets, whereas the others are primarily short baseline. We observe larger gains on the short-baseline datasets; under long baselines, stronger perspective changes and scale/deformation effects undermine the assumptions underlying our geometric-consistency term, leading to more limited improvements. This behavior aligns with the method’s intended operating regime.

       \subsubsection{Runtime Analysis} Table~\ref{tab:runtime} summarizes the average runtime efficiency on the Fundamental Matrix (F-Matrix), Relative Pose (Rel. Pose), and Homography (Homo.) tasks. Although standard RANSAC~\cite{fischler1981random} and LO-RANSAC~\cite{chum2003locally} exhibit lower latency due to their algorithmic simplicity, this comes at the cost of reduced estimation accuracy, as evidenced in Table I. In comparison to the state-of-the-art GC-RANSAC~\cite{GC_RANSAC}, which necessitates computationally intensive graph-cut optimization, our DLOSAC achieves an approximate 39\% reduction in runtime. While the primary computational overhead in DLOSAC stems from Delaunay triangulation, this process is performed solely on the subset of candidate inliers rather than the entire correspondence set. Consequently, the computational cost remains tractable, supporting the method's applicability to real-time robotic systems.
 
       \input{table/table_runtime}

    \subsection{Ablation study}
    \label{sec:ablation}
        \input{table/table2}
        DLOSAC consists of two regularization terms: an edge-based geometric consistency term ($\mathcal{L}_{\text{edge}}$) and a spatial distribution term ($\mathcal{L}_{\text{spatial}}$). The ablation study aims to assess the contribution of each of these terms. Table~\ref{tab:table2} presents the results of the ablation experiments across three tasks (1st row). By progressively removing the regularization terms of DLOSAC, we evaluate the cross-dataset performance of each variant across the three tasks. The last row shows the performance of the full DLOSAC model, with the effectiveness of each term validated by separately removing the spatial distribution term (w/o $\mathcal{L}_{\text{spatial}}$) and the edge-based geometric consistency term (w/o $\mathcal{L}_{\text{edge}}$). Finally, we also remove both regularization terms (w/o $\mathcal{L}_{\text{edge}}+\mathcal{L}_{\text{spatial}}$) to demonstrate the performance of the baseline method.

    \subsection{Sensitivity Analyses}
    \label{sec:sensitivity}

\begin{figure}[t!]
            \centering
            \begin{subfigure}[t]{0.481\linewidth}
                \includegraphics[width=\linewidth]{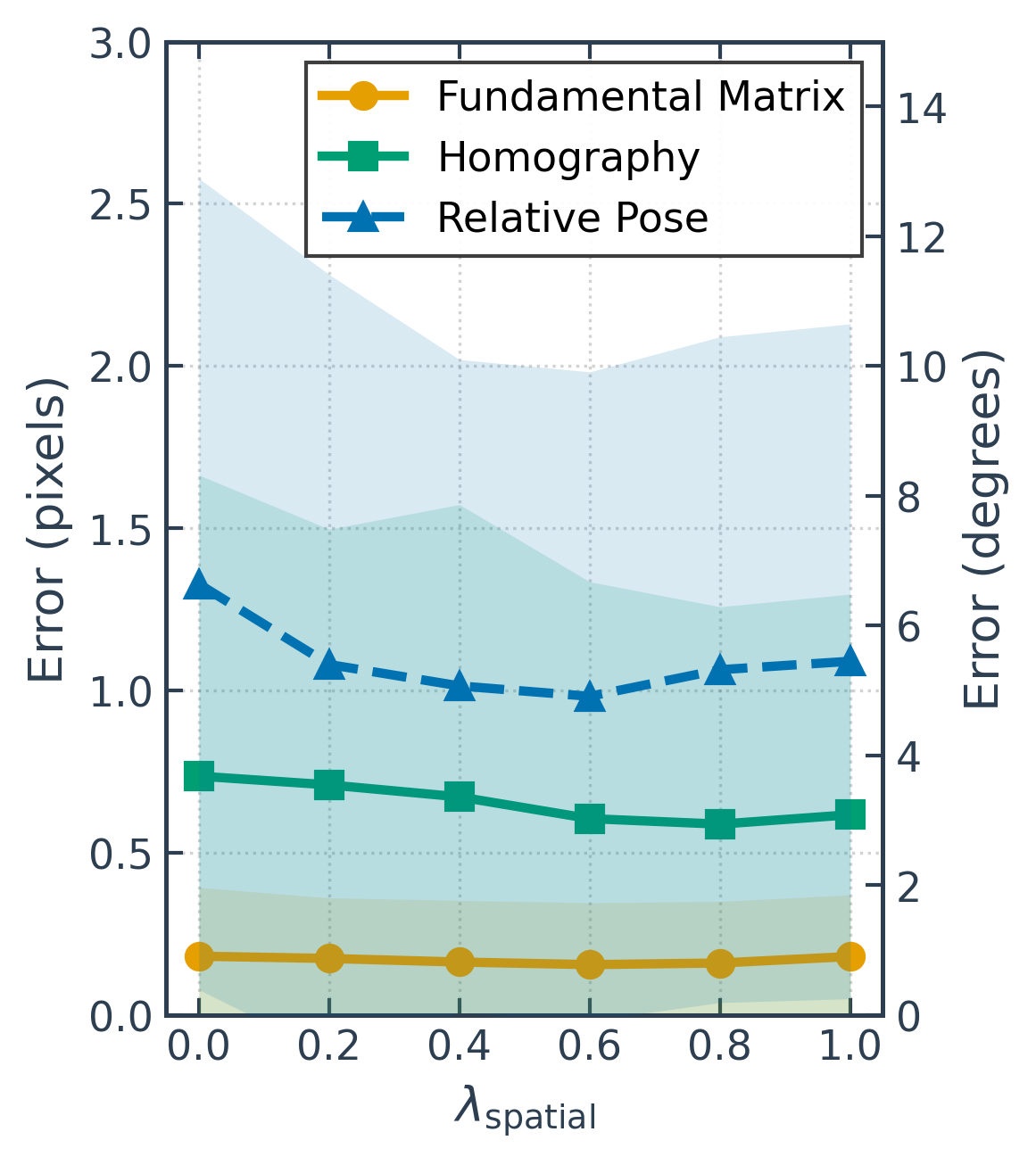}
                \caption{}
            \end{subfigure}
            \begin{subfigure}[t]{0.481\linewidth}
                \includegraphics[width=\linewidth]{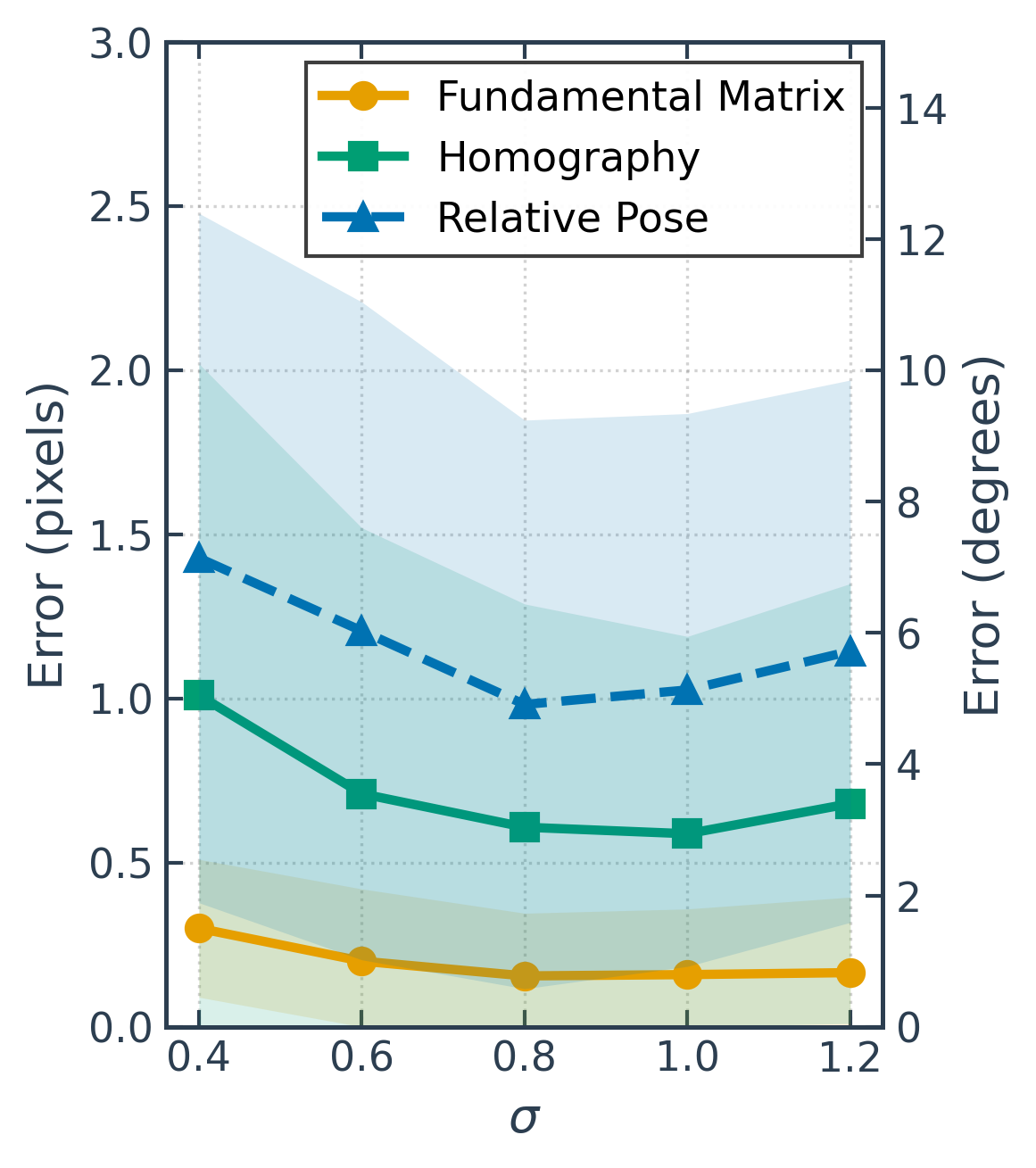}
                \caption{}
                \label{subfig:s2}
            \end{subfigure}
            
            \caption{Parameter analysis: (a) Weighting coefficient for the spatial distribution loss \( \lambda_{\text{spatial}} \); (b) Regularization weight \( \sigma \).}

            \label{fig:sensitivity}
        \end{figure}

        DLOSAC controls the contribution of the spatial-distribution loss via the hyperparameter $\lambda_{\text{spatial}}$. Setting this weight too large can weaken geometric consistency, while setting it too small tends to produce clustered keypoints and local overfitting. To select an appropriate value, we performed a parameter sweep over $[0,1]$ and evaluated cross-dataset average errors on three tasks. Fig.~\ref{fig:sensitivity} summarizes the results: the left axis reports the average pixel error $\varepsilon_{\text{avg}}$ for fundamental matrix and homography estimation, while the right axis reports the average rotation and translation errors for relative-pose estimation. The curves reach a low-error region around $0.6$--$0.8$; therefore, we set $\lambda_{\text{spatial}}=0.6$ for all experiments. Moreover, the error curves vary smoothly within this range, indicating low sensitivity to this hyperparameter.

        The parameter $\sigma$ controls the sharpness of the exponential weighting function. Smaller $\sigma$ amplifies differences in the weights and penalizes high-loss points more strongly, whereas larger $\sigma$ smooths the weights and weakens the regularization effect. To identify an appropriate range, Fig.~\ref{fig:sensitivity}(\subref{subfig:s2}) reports cross-dataset average errors for three geometric estimation tasks under varying $\sigma$. The left axis shows the average pixel error for fundamental-matrix and homography estimation, while the right axis shows the average rotation and translation errors for relative-pose estimation. The curves remain smooth and reach a stable low-error region around $\sigma=0.8$; therefore, we adopt $\sigma=0.8$ as the default value, balancing stability and performance across all three tasks.

%% file: table/table1.tex
\begin{table*}[t!]
\centering
\caption{The error and improvement rates for all methods (1st row) across tasks (1st column) and datasets (2nd column) are presented, along with the average performance (Avg.) of each task across benchmarks. The optimal results are indicated in \textbf{bold}, and the second-best results are marked with \underline{underline}.}
\label{tab:t1}
\renewcommand{\arraystretch}{1.05}
\resizebox{\textwidth}{!}{
\begin{tabular}{c c l c c c c c c}
\toprule
 &  &  & RANSAC~\cite{fischler1981random} & LO-RANSAC~\cite{chum2003locally} & $\text{LO}^{+}$-$\text{RANSAC}$~\cite{LO+RANSAC}  & GC-RANSAC~\cite{GC_RANSAC} & PSC-RANSAC~\cite{PSC_RANSAC} & DLOSAC (ours) \\
\midrule

\multirow{15}{*}{\rotatebox{90}{\textbf{Fundamental Matrix}}}

 & \multirow{3}{*}{\rotatebox{90}{TUM}}
   & $\epsilon_{\text{avg}}$ (px) & 0.482 & 0.294 & 0.247 & \underline{0.139} & 0.148 & \textbf{0.116} \\
 &  & $\epsilon_{\text{std}}$ (px) & 0.478 & 0.230 & 0.225 & 0.193 & \underline{0.190} & \textbf{0.130} \\
 &  & $f_{imp}$ (\%) &  75.93&  60.54&  53.04&  \underline{16.55} &  21.62&  \textbf{0.00}\\
\cmidrule(lr){3-9}

 & \multirow{3}{*}{\rotatebox{90}{T\&T}}
  & $\epsilon_{\text{avg}}$ (px) & 0.538 & 0.402 & 0.329 & 0.229 & \underline{0.204} & \textbf{0.197}\\
 &  & $\epsilon_{\text{std}}$ (px) & 0.501 & 0.398 & 0.394 & 0.263 & \underline{0.250} & \textbf{0.214}\\
 &  & $f_{imp}$ (\%) &63.38  &51.00  &40.12  &13.97  &\underline{3.43}  &\textbf{0.00}  \\
\cmidrule(lr){3-9}

 & \multirow{3}{*}{\rotatebox{90}{KITTI}}
  & $\epsilon_{\text{avg}}$ (px) & 0.493& 0.338 & 0.310 & \underline{0.134}& 0.154 & \textbf{0.115} \\
 &  & $\epsilon_{\text{std}}$ (px) & 0.582 & 0.351 & 0.307 & \underline{0.201} & \textbf{0.193} & 0.203 \\
 &  & $f_{imp}$ (\%) &  76.67&  65.98&  62.90&  \underline{14.18}&  25.32&  \textbf{0.00}\\
\cmidrule(lr){3-9}

 & \multirow{3}{*}{\rotatebox{90}{CPC}}
  & $\epsilon_{\text{avg}}$ (px) & 0.777 & 0.573 & 0.489 & 0.225 & \underline{0.210} & \textbf{0.194} \\
 &  & $\epsilon_{\text{std}}$ (px) & 0.757 & 0.501 & 0.511 & 0.313 & \underline{0.235} & \textbf{0.216} \\
 &  & $f_{imp}$ (\%) &  75.03&  66.14&  60.33&  13.78&  \underline{7.62}&  \textbf{0.00}\\
\cmidrule(lr){3-9}


 & \multirow{3}{*}{\rotatebox{90}{Avg.}}
 & $\epsilon_{\text{avg}}$ (px) & 0.573 & 0.402 & 0.344 & 0.182 & \underline{0.179} & \textbf{0.156} \\
 &  & $\epsilon_{\text{std}}$ (px) & 0.580 & 0.370 & 0.359 & 0.243 & \underline{0.217} & \textbf{0.191} \\
 &  & $f_{imp}$ (\%) & 72.77 & 61.19 & 54.65 & 14.29 & \underline{12.85} & \textbf{0.00} \\

\midrule

\multirow{15}{*}{\rotatebox{90}{\textbf{Relative Pose}}}
 & \multirow{3}{*}{\rotatebox{90}{TUM}}
  & $\epsilon_R$ (°) & 1.247 & 0.924 & 0.881 & \underline{0.498} & 0.563  &  \textbf{0.471} \\
 &  & $\epsilon_t$ (°) & 14.396 & 13.964 & 13.151 & \underline{13.149} & 13.217 & \textbf{9.757} \\
 &  & $f_{\text{imp}}$ (\%) &  34.62& 31.30& 27.11&  \underline{25.05}&  25.78&  \textbf{0.00}\\
\cmidrule(lr){3-9}

 & \multirow{3}{*}{\rotatebox{90}{T\&T}}
  & $\epsilon_R$ (°) & 5.195 & 4.786 & 3.634 & 3.218 & \underline{3.089}  & \textbf{2.910} \\
 &  & $\epsilon_t$ (°) & 11.478 & 9.646 & 9.064 & 8.605 & \underline{7.338}  & \textbf{6.630} \\
 &  & $f_{\text{imp}}$ (\%) &  42.78&  33.90&  24.87&   19.31& \underline{8.51}&  \textbf{0.00}\\
\cmidrule(lr){3-9}

 & \multirow{3}{*}{\rotatebox{90}{KITTI}}
  & $\epsilon_R$ (°) & 1.104 & 0.979 & 0.905 & \underline{0.643} & 0.752 & \textbf{0.590} \\
 &  & $\epsilon_t$ (°) & 7.522 & 5.546 & 5.515 & 4.956 & \underline{4.806} & \textbf{4.149}\\
 &  & $f_{\text{imp}}$ (\%) &  45.06&  27.37&  26.18&  15.36& \underline{14.74}&  \textbf{0.00}\\
\cmidrule(lr){3-9}

 & \multirow{3}{*}{\rotatebox{90}{CPC}}
  & $\epsilon_R$ (°) & 8.985 & 6.642 & 6.192 & 5.541 &\textbf{5.216} & \underline{5.307} \\
 &  & $\epsilon_t$ (°) & 18.846 & 13.690 & 13.715 & 13.406 & \underline{10.618} & \textbf{9.469} \\
 &  & $f_{\text{imp}}$ (\%) &  46.91&  27.33&  25.77&  22.01&  \underline{6.68}& \textbf{0.00}  \\
 \cmidrule(lr){3-9}

  & \multirow{3}{*}{\rotatebox{90}{Avg.}}
  & $\epsilon_R$ (°) & 4.133 & 3.333 & 2.903 & 2.475 & \underline{2.405} & \textbf{2.320} \\
 &  & $\epsilon_t$ (°) & 13.061 & 10.712 & 10.361 & 10.029 & \underline{8.995} & \textbf{7.501} \\
 &  & $f_{\text{imp}}$ (\%) & 42.88& 30.07& 25.96& 21.46& \underline{13.85}& \textbf{0.00} \\

\midrule

\multirow{9}{*}{\rotatebox{90}{\textbf{Homography}}}
 & \multirow{3}{*}{\rotatebox{90}{Homo}}
  & $\epsilon_{\text{avg}}$ (px) & 0.920 & 0.521 & 0.503 & 0.349 & \underline{0.287} & \textbf{0.246} \\
 &  & $\epsilon_{\text{std}}$ (px) & 0.810 & 0.833 & 0.784 & 0.631 & \underline{0.569} & \textbf{0.537} \\
 &  & $f_{\text{imp}}$ (\%) &  73.26&  52.78&  51.09&  29.51&  \underline{14.29}&  \textbf{0.00}\\
\cmidrule(lr){3-9}

 & \multirow{3}{*}{\rotatebox{90}{EVD}}
  & $\epsilon_{\text{avg}}$ (px) & 1.973 & 1.496 & 1.358 & \underline{1.049} & 1.187 & \textbf{0.929} \\
 &  & $\epsilon_{\text{std}}$ (px) & 2.012 & 1.611 & 1.629 & \underline{1.367} & 1.480 & \textbf{1.266} \\
 &  & $f_{\text{imp}}$ (\%) &  52.91&  37.90&  31.59&  \underline{11.44}&  21.74&  \textbf{0.00}\\
\cmidrule(lr){3-9}

 & \multirow{3}{*}{\rotatebox{90}{Avg.}}
  & $\epsilon_{\text{avg}}$ (px) & 1.447 & 1.009 & 0.931 & \underline{0.699} & 0.737 & \textbf{0.588} \\
 &  & $\epsilon_{\text{std}}$ (px) & 1.411 & 1.222 & 1.207 & \underline{0.999} & 1.025 & \textbf{0.902} \\
 &  & $f_{\text{imp}}$ (\%) & 59.36 & 41.72 & 36.84 & \underline{15.88} & 20.22 & \textbf{0.00} \\

\bottomrule
\end{tabular}
}
\end{table*}

%% file: table/table_runtime.tex
\begin{table}[t!]
\centering
\caption{Comparison of average runtime (ms). Our method is highlighted in \colorbox{lightblue}{blue}. \textbf{Bold} and \underline{underlined} values indicate the best and second-best performance, respectively.}
\label{tab:runtime}
\renewcommand{\arraystretch}{1.0}
\setlength{\tabcolsep}{8pt}
\resizebox{\columnwidth}{!}{%
\begin{tabular}{lcccc}
\toprule
\textbf{Method} & \textbf{F-Matrix} & \textbf{Rel. Pose} & \textbf{Homo.} & \textbf{Avg.} \\
\midrule
RANSAC      & \textbf{18.24} & \textbf{24.62} & \textbf{8.26}  & \textbf{17.04} \\
LO-RANSAC      & 29.66          & 41.18          & 16.41          & 29.08 \\
LO$^+$-RANSAC      & \underline{26.43} & \underline{38.08} & \underline{14.24} & \underline{26.25} \\
GC-RANSAC           & 65.32          & 92.56          & 45.11          & 67.66 \\
PSC-RANSAC          & 72.17          & 98.32          & 48.22          & 72.90 \\
\midrule
\rowcolor{lightblue}
DLOSAC (Ours)                         & 42.65          & 58.90          & 22.73          & 41.43 \\
\bottomrule
\end{tabular}
}
\end{table}

%% file: table/table2.tex
\begin{table}[t!]
  \centering
  \caption{\textbf{Ablation Study.} Mean errors and standard deviations across all datasets for three geometric estimation tasks. For relative-pose estimation, rotation and translation errors are reported separately. The best results are highlighted in \textbf{bold}.}
  \label{tab:table2}
  \renewcommand{\arraystretch}{1.1}
  \setlength{\tabcolsep}{3pt}
  \resizebox{\columnwidth}{!}{%
  \large
  \begin{tabular}{l cc cc cc}
    \toprule
    \multirow{2.5}{*}{\textbf{Method}} & \multicolumn{2}{c}{\textbf{Fund. Matrix}} & \multicolumn{2}{c}{\textbf{Rel. Pose}} & \multicolumn{2}{c}{\textbf{Homo.}} \\
    \cmidrule(lr){2-3} \cmidrule(lr){4-5} \cmidrule(lr){6-7}
     & $\epsilon_{\text{avg}}$ (px) & $\epsilon_{\text{std}}$ (px) 
     & $\epsilon_{\text{R}}$ ($^\circ$)& $\epsilon_{\text{t}}$ ($^\circ$)& $\epsilon_{\text{avg}}$ (px) & $\epsilon_{\text{std}}$ (px) \\
    \midrule
    w/o ($\mathcal{L}_{\text{edge}}+\mathcal{L}_{\text{spatial}}$)
     & 0.573 & 0.580 & 4.133 & 13.061 & 1.447 & 1.411 \\
    w/o ($\mathcal{L}_{\text{edge}}$)  
    & 0.295 & 0.371 & 2.755 & 9.122 & 0.935 & 1.173 \\
    w/o ($\mathcal{L}_{\text{spatial}}$) 
    & 0.181 & 0.212 & 2.643 & 10.90 & 0.736 & 0.928 \\
    \midrule
    \rowcolor{lightblue}
    DLOSAC (Ours) 
    & \textbf{0.156} & \textbf{0.191} & \textbf{2.320} & \textbf{7.501} & \textbf{0.588} & \textbf{0.902} \\
    \bottomrule
  \end{tabular}%
  }
\end{table}

%% file: sec/Conclusion.tex
\section{Conclusion}
    In this paper, we introduced DLOSAC, a novel local optimization framework that integrates both geometric consistency and spatial distribution information to improve geometric transformation estimation. Unlike existing methods that mainly focus on point-level geometric constraints, DLOSAC takes advantage of the spatial distribution of inliers by incorporating Delaunay triangulation, edge-level consistency metrics, and distribution descriptors. These innovations lead to more robust inlier selection and model refinement. Extensive experiments across six public datasets and three geometric estimation tasks demonstrate that our method consistently achieves superior performance compared to existing state-of-the-art methods. Ablation studies and sensitivity analysis further confirm the effectiveness and robustness of the joint geometric consistency and spatial distribution optimization strategy.

    Although DLOSAC outperforms baseline methods across six public benchmarks, its gains are relatively modest in wide-baseline settings (e.g., CPC and T\&T). The primary cause of this limitation is that the current geometric regularizers rely heavily on the assumption of approximate Euclidean invariance. When significant viewpoint or scale changes occur, especially with strong parallax, pronounced depth variation, and non-coplanar structures, this assumption breaks down. As a result, true-inlier residuals are systematically inflated, reducing inlier–outlier separability and weakening mismatch discrimination. Notably, this issue is not unique to our approach; algorithms incorporating spatial-level constraints, such as GC-RANSAC~\cite{GC_RANSAC} and PSC-RANSAC~\cite{PSC_RANSAC}, are similarly affected.
    
    In future work, we will (a) develop projective- and affine-invariant regularizers with epipolar-based reweighting to maintain stability under large viewpoint changes and strong parallax; (b) employ multi-scale triangulation and distributional matching using a coarse-to-fine schedule with adaptive weighting of node- and distribution-level terms to mitigate bias arising from unstable local statistics; and (c) integrate DLOSAC’s local optimization into a complete, state-of-the-art RANSAC framework (e.g., USACv20~\cite{USACv20}).